\pdfoutput=1

\documentclass[11pt]{article}

\usepackage{acl}

\usepackage{times}
\usepackage{latexsym}

\usepackage[T1]{fontenc}

\usepackage[utf8]{inputenc}

\usepackage{microtype}

\usepackage{xspace}
\usepackage{multirow}
\usepackage{booktabs}
\usepackage{xcolor}
\usepackage{caption}
\usepackage{graphicx}
\usepackage[frozencache,cachedir=minted-cache]{minted}
\usepackage{soul}
\usepackage{amssymb}

\usepackage{amsmath,amsfonts,bm}

\def\eqref#1{equation~\ref{#1}}

\def\1{\bm{1}}

\def\ve{{\bm{e}}}

\def\vq{{\bm{q}}}

\def\vs{{\bm{s}}}

\def\vx{{\bm{x}}}
\def\vy{{\bm{y}}}

\DeclareMathAlphabet{\mathsfit}{\encodingdefault}{\sfdefault}{m}{sl}
\SetMathAlphabet{\mathsfit}{bold}{\encodingdefault}{\sfdefault}{bx}{n}

\newcommand{\methodabbr}{QueryForm\xspace}
\newcommand{\dprompt}{S-Prompt\xspace}
\newcommand{\eprompt}{E-Prompt\xspace}
\newcommand{\pagemage}{QueryWeb\xspace}
\newcommand{\pagemageabbr}{QW\xspace}
\newcommand{\biprompt}{dual prompting\xspace}

\newcommand{\ie}{\textit{i.e.}}
\newcommand{\eg}{\textit{e.g.}}

\title{\methodabbr: A Simple Zero-shot Form Entity Query Framework}

\author{Zifeng Wang\textsuperscript{1}\thanks{\ Work done while the author was an intern at Google Cloud AI Research. Email:~{zifengwang@ece.neu.edu}}, Zizhao Zhang\textsuperscript{2}$^\dagger$, Jacob Devlin\textsuperscript{3}, Chen-Yu Lee\textsuperscript{2}, \\ \bf Guolong Su\textsuperscript{3}, Hao Zhang\textsuperscript{3}, Jennifer Dy\textsuperscript{1}, Vincent Perot\textsuperscript{3}$^\dagger$, and Tomas Pfister\textsuperscript{2} \\
\textsuperscript{1}Northeastern University \quad \textsuperscript{2}Google Cloud AI \quad  \textsuperscript{3}Google Research\\
\small{$^\dagger$Core contribution} 
}

\begin{document}
\maketitle
\begin{abstract}
Zero-shot transfer learning for document understanding is a crucial yet under-investigated scenario to help reduce the high cost involved in annotating document entities. 
We present a novel query-based framework, \methodabbr, that extracts entity values from form-like documents in a zero-shot fashion. 
\methodabbr contains a \biprompt mechanism that composes both the document schema and a specific entity type into a query, which is used to prompt a Transformer model to perform a single entity extraction task.
Furthermore, we propose to leverage large-scale query-entity pairs generated from form-like webpages with weak HTML annotations to pre-train \methodabbr. By unifying pre-training and fine-tuning into the same query-based framework, \methodabbr enables models to learn from structured documents containing various entities and layouts, leading to better generalization to target document types without the need for target-specific training data.~\methodabbr sets new state-of-the-art average F1 score on both the XFUND (+4.6\%$\sim$10.1\%) and the Payment (+3.2\%$\sim$9.5\%) zero-shot benchmark, with a smaller model size and no additional image input. 
\end{abstract}

\section{Introduction}

Form-like document understanding has become a booming research topic recently thanks to its many real-world applications in industry. 
Form-like documents refer to documents with rich typesetting formats, such as invoices and receipts in everyday inventory workflow. 
Automatically extracting and organizing structured information from form-like documents is a valuable yet challenging problem.

Recent methods~\citep{xu2020layoutlm, garncarek2021lambert, lee2022formnet} often discuss the problem of form-like document understanding, e.g.~document entity extraction (DEE), in the supervised setting, assuming the training and test sets are of the same document type. 
However, in real-world scenarios, there is often the need for generalizing models from seen document types to new unseen document types. 
Beyond annotation costs, endlessly training specialized models on new types of documents is not scalable in many practical scenarios. 
Moreover, the methods in the supervised setting pre-define the document schema, \ie the set of entities contained in the document, following the sequence-to-sequence tagging framework via the BOISE labeling format~\citep{ratinov2009design}. 
Consequently, the models lack the ability to learn from different documents with diverse schemas.

\begin{figure}
    \centering
    \includegraphics[width=0.99\linewidth]{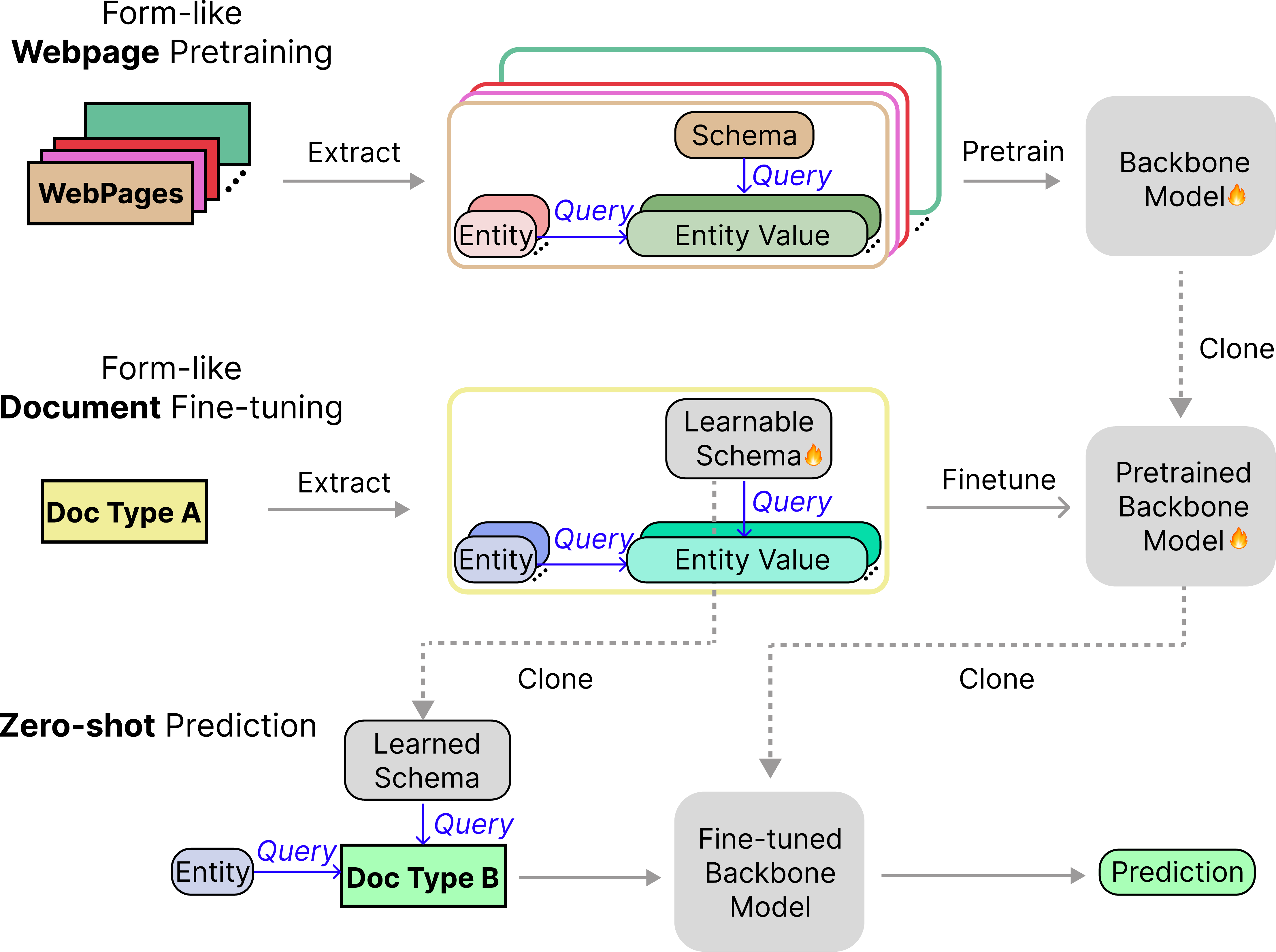}
    \caption{Illustration of the zero-shot transfer learning stages of \methodabbr. In the pre-training stage, we extract millions of schemas and entity-value pairs from publicly available webpages to generate a large amount of query-value pairs to teach the backbone model to make query-conditional prediction. 
    During fine-tuning, we extract more accurate entity-value pairs from the available annotated document and directly learn schema information from data. 
    Finally, we evaluate the pre-trained model on a different target document without training data.}
    \label{fig:intro}
    \vspace{-4mm}
\end{figure}

\begin{figure}[t]
    \centering
    \includegraphics[width=0.999\linewidth]{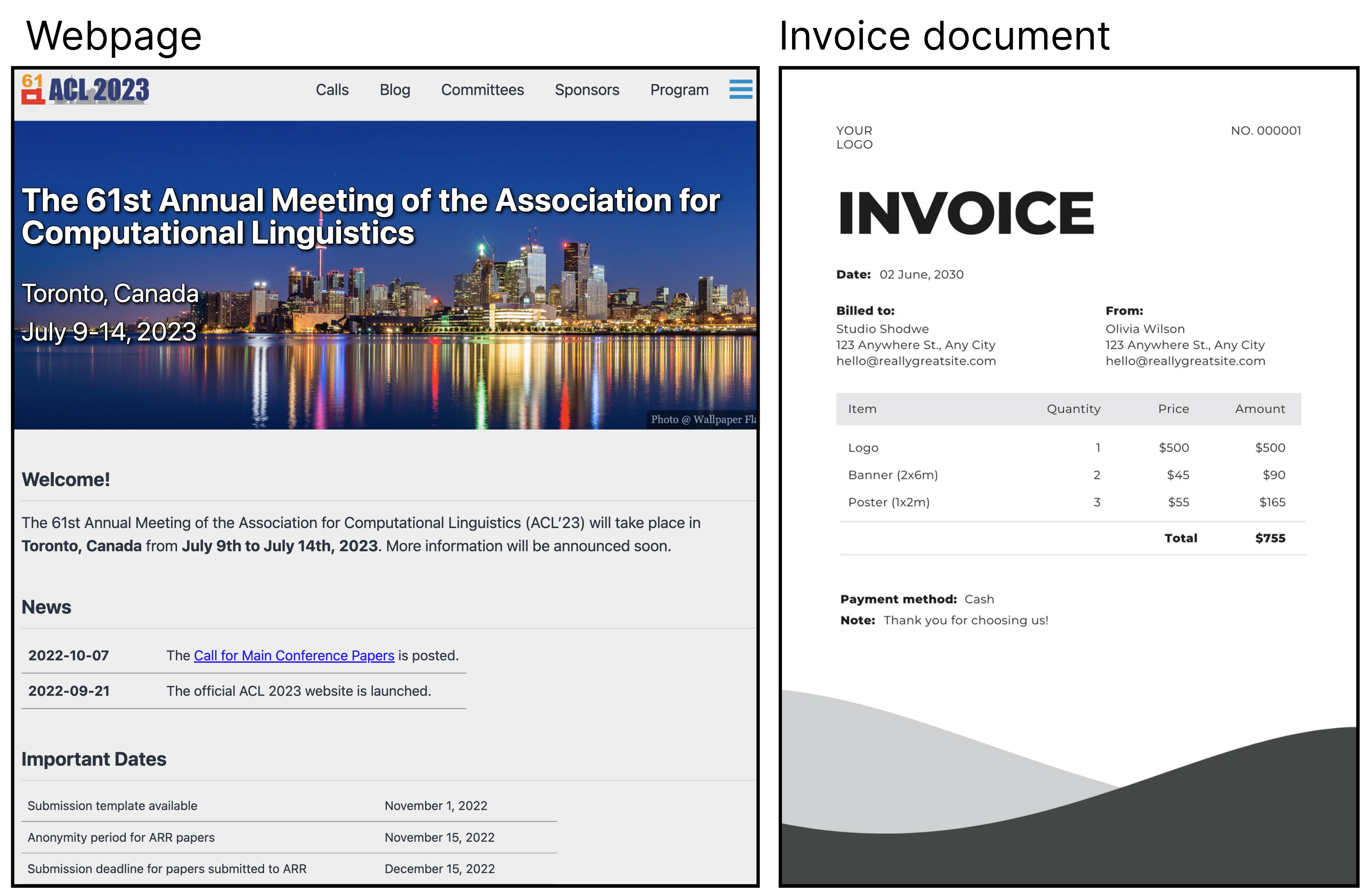}
    \caption{Form-like examples of Webpage and Invoice documents. Webpage appears to have distinct layouts and contents from invoice documents, but they both contain rich entity-value pairs, such as ``\textit{page title}-The 61st Annual Meeting of the Association for Computational Linguistics'' in Webpage and ``\textit{total amount} - \$755'' in Invoice.}
    \label{fig:po_vs_invoice}
\end{figure}

Thus, it is desirable to have a systematic way to learn knowledge from existing annotated documents of different types to an un-annotated target document type (e.g. invoice in Figure~\ref{fig:po_vs_invoice}, right). This learning paradigm is usually defined as zero-shot transfer learning in literature~\cite{xu2021layoutxlm}. Beyond this, it is even more desirable to leverage highly-structured form-like documents with rich schema, such as form-like webpages in Figure~\ref{fig:po_vs_invoice}, left. Although webpages do not have explicit human annotations, we believe the diverse schemas and natural ``entities'', such as headers and text paragraphs, that exist in webpages can be valuable for document understanding. However, how to effectively utilize these webpages with a high discrepancy from documents like invoice and receipt is an unknown yet challenging problem.

In this work, we propose a novel query-based framework, \methodabbr, to learn transferable knowledge from different types of documents for zero-shot entity extraction on the target document type. The workflow of \methodabbr is illustrated in Figure~\ref{fig:intro}. Ideally, we would like to prompt the model: \textit{This document has the following \textsc{[schema]}, please extract its \textsc{[entity]} value}, and model is able to accurately predict the corresponding word tokens belong to the queried entity. %
To this end, we encode both schema and entity information in our query, so that the model is no longer limited by a certain document type and a fixed set of entity types (or classes). Moreover, our query-based design can even benefit further from large-scale datasets with diverse schemas and entity types.

In order to feed this kind of composite query, we propose a \textit{\biprompt} strategy to effectively prompt the backbone model, \eg, a pre-trained Transformer, to make conditional prediction. As its name suggests, the \biprompt strategy consists of an E(ntity)-Prompt and a S(chema)-Prompt. Depending on the annotations we have, we can either generate the prompts from semantic labels, or learn them directly from data. Although similar concepts to \biprompt exist in the vision field~\citep{wang2022dualprompt,wang2022learning} to solve different problems, the main design in \methodabbr is original in DEE.
We also propose a query-based pre-training method, \pagemage, which leverages the highly-accessible and inexhaustible resource - publicly available webpages.
During the pre-training stage, the model learns to quickly adapt to various queries composed of different \dprompt{s} and \eprompt{s} generated from HTML source of webpages. After the decoupling of entity and schema that are tied to document types, the model can learn more transferable knowledge - leveraging the rich layout, scale and content information in webpages to make query-conditional predictions.

In summary, our work makes the following contributions:
\begin{itemize}
    \item We propose \methodabbr, a novel yet simple query-based framework for zero-shot document entity extraction. \methodabbr provides a new \biprompt mechanism to encode both document schema and entity information to learn transferrable knowledge from source to target document types.
    \item %
    We demonstrate an effective pre-training approach, \pagemage, that collects publicly available webpages with various layouts and HTML sources, and pre-trains \methodabbr via the \biprompt mechanism.
    Although webpages show high discrepancy from the target documents, we show this approach consistently improves the zero-shot performance.
    \item With extensive empirical evaluation, \methodabbr sets new state-of-the-art F1 score on both Inventory-Payment and FUNSD-XFUND zero-shot transfer learning benchmarks.
    
\end{itemize}

\section{Related Work}

\textbf{Document entity extraction (DEE).} Researchers started to study extracting information from documents using rule-based models \citep{lebourgeois1992fast,o1993document,simon1997fast}, or learning-based approaches with hand-engineered features~\cite{marinai2005artificial,wei2013evaluation,schuster2013intellix}. These methods have limited representation power and generalization ability.

\noindent More recently, neural models have been the mainstream solution for document entity extraction (DEE). Both RNN-based~\citep{palm2017cloudscan,aggarwal2021form2seq} and CNN-based models~\citep{katti2018chargrid,zhao2019cutie,denk2019bertgrid} have been adopted for DEE task. Nevertheless, motivated by the superior performance of the Transformers~\citep{vaswani2017attention} in various NLU tasks~\citep{devlin2018bert,raffel2020exploring}, recent work develops multiple Transformer-based models for DEE. \citet{majumder2020representation} extended BERT~\cite{devlin2018bert} to learn representations for form-like documents;
~\cite{kim2022ocr} propose a encoder-decoder structure that directly extracts document information from image input; ~\citet{xu2020layoutlm,xu2021layoutxlm} leverage both image and text inputs to capture cross-modality information.~\citet{lee2021rope,lee2022formnet} further introduced GCN~\cite{kipf2016semi} to encode spatial relationships in addition to the Transformer backbone. However, these methods only consider the usual supervised learning setting, \ie, training and test sets are from the same document type. On the contrary, our work proposes a novel query-based framework and tackle the challenging yet under-investigated zero-shot transfer learning~\citep{xu2021layoutxlm} setting.

\noindent \textbf{Pre-training for DEE.} Existing large-scale pre-training techniques in NLP~\cite{devlin2018bert,conneau2019unsupervised,liu2019roberta} are readily available for serialized document tokens. Multimodal pre-training~\citep{xu2020layoutlmv2,xu2020layoutlm,xu2021layoutxlm,appalaraju2021docformer} achieves better performance than text-modality alone by incorporating visual information at the cost of more expensive data collection and computation costs. 
Our work presents a novel pre-training method using text modality alone, which is complementary to models that relies on image modality~\cite{kim2022ocr} or multiple modalities~\cite{xu2020layoutlm, xu2021layoutxlm}.
Moreover, we leverage publicly available webpages, which contain rich structured information and are much more accessible than documents. Different from the common Mask Language Model (MLM) objective used in pre-training, \methodabbr has the same query-conditional objective during both pre-training and fine-tuning, which intuitively strengthens the transferability of pre-trained knowledge.

To the best of our knowledge, DQN~\citep{gao2022docquerynet} and Donut~\citep{kim2022ocr} are the closest work to ours in the DEE domain. However, our work is still very different from them from multiple perspectives, including problem setting, query design, and pre-training technique. On the other hand, leveraging webpages to pre-train language models has been explored in prior work.~\citet{liu2019roberta, brown2020language} extract text corpora from webpages and~\cite{aghajanyan2021htlm} use HTML source for pre-training. However, to the best of our knowledge, we are the first to leverage both webpages and the corresponding HTML source in a novel query-based pre-training framework to address the challenging zero-shot DEE task. Our framework fully takes advantage of the rich schema and layout information from webpages and utilizes HTML tags as weak entity annotation to align pre-training with the downstream DEE task.

\begin{figure*}[t]
    \centering
    \includegraphics[width=0.95\linewidth]{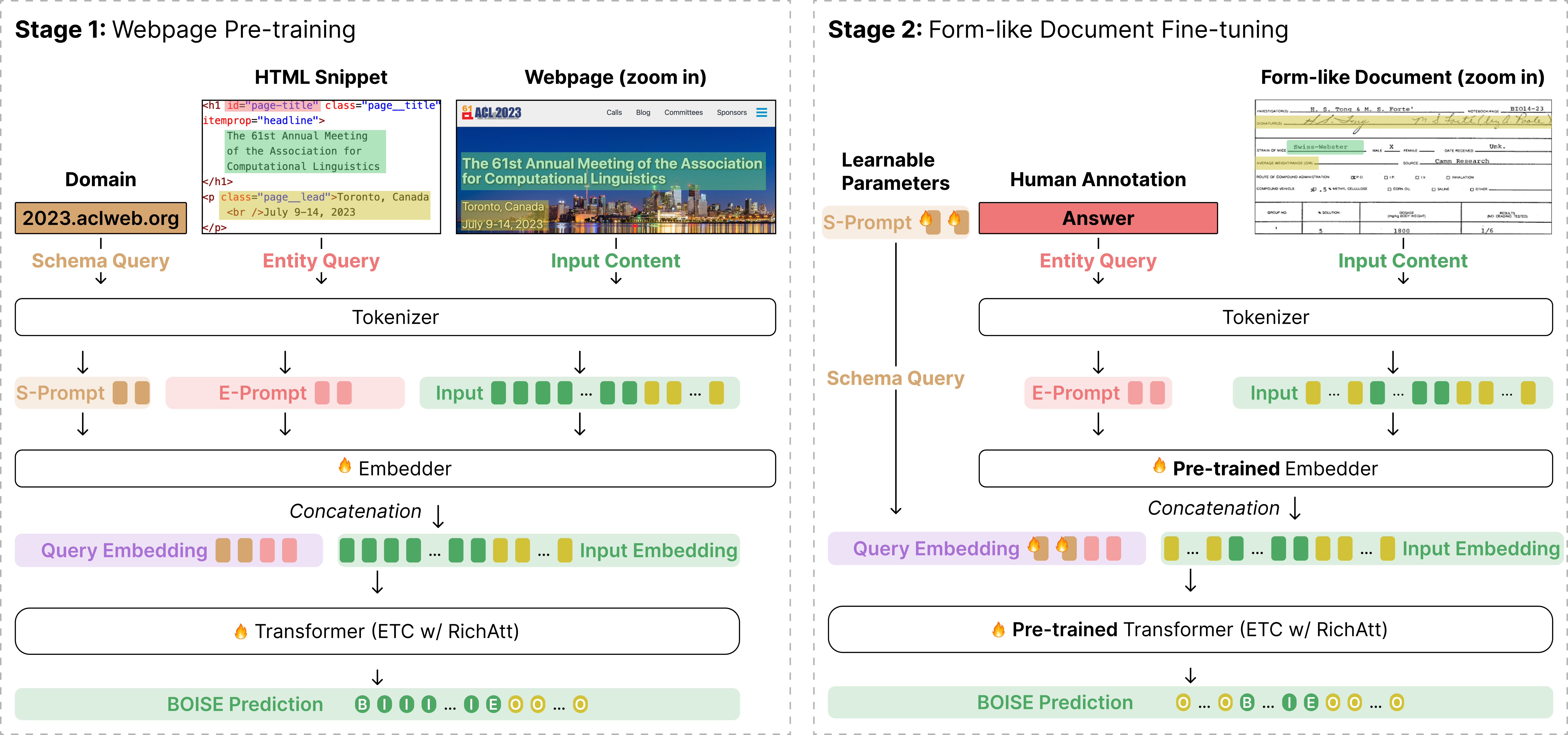}
    \caption{Overview of \methodabbr. Our \biprompt design yields a consistent objective in both pre-training and fine-tuning stages. Note that the schema query in pre-training comes from website domains while it is a learnable parameter in fine-tuning. See Section~\ref{sec:method} for more details.}
    \label{fig:overview}
\end{figure*}

\section{Preliminaries} 
\subsection{Problem Formulation} \label{sec:problem_formulation}
Given serialized words from a form-like document, we formulate the DEE problem as sequence tagging for tokenized words, \ie, for each word, we predict its corresponding entity class. Recent methods~\citep{xu2020layoutlm, garncarek2021lambert, lee2022formnet} use the BOISE labeling format~\citep{ratinov2009design} - classifying the token as \{$\ve$-Begin, Outside, $\ve$-Inside, $\ve$-Single, $\ve$-End\} of a certain entity $\ve \in \mathbf{E}$ to mark the entity span, where $\mathbf{E}$ is the set of entities of interest. Thus the cardinality of the label space will be $(4 \times |\mathbf{E}| + 1)$. In our formulation, we explicitly encode entity in the \eprompt. Therefore, we are able to use a more succinct and generalizable BOISE labeling with only $5$ labels, \{Begin, Outside, Inside, Single, End\} to mark the span. Our approach decouples the label space with entity types. Following~\citet{lee2022formnet}, we then apply the Viterbi algorithm to get the final prediction.

In our work, we focus on the zero-shot DEE setting proposed by~\citet{xu2021layoutxlm}, where 1) the training source documents have significant domain gap from the target test documents (e.g. languages or document types), 2) there is no training documents available from the target documents, and 3) source documents include entities contained in the target documents.

\subsection{Architecture design} \label{sec:arch_design}
Following the setting in earlier work~\citep{majumder2020representation, lee2022formnet}, our method takes the WordPiece~\cite{wu2016google} tokenized outputs from the Optical Character Recognition (OCR) engine in reading order (left-right and top-bottom). By design, our method is compatible with any sequence encoder model as the backbone. We adopt the long-sequence transformer extension, ETC~\citep{ainslie2020etc} as our backbone, following the adoption of~\citet{lee2022formnet}, which contains Rich Attention as an enhancement of self-attention layers to encode 2D spatial layout information.  We find this method (used as our baseline) performs fairly strong in the usual supervised learning setup, however, its performance drops significantly in the zero-shot learning setting. 

Note that in practice, one can use \methodabbr with OCR engines with different heuristics or other model backbones~\cite{zaheer2020big}. The work focuses on how to enrich entity query abilities from forms via our proposed \methodabbr.

\section{Methodology} \label{sec:method}

We propose~\methodabbr as a general query-based framework for solving the zero-shot DEE problem. \methodabbr consists of a novel \biprompt strategy and a specially-designed pre-training approach called \pagemage. In Figure~\ref{fig:overview}, the model is first pre-trained on a large-scale webpage dataset to learn to make conditional prediction under relatively noisy queries generated from combination of webpage domains (proxy of schema) and HTML tags (proxy of entity). Then, the model is fine-tuned on form-like documents with a unified schema to learn more specialized knowledge, by learning schema information in \dprompt and further encode more accurate entity-level knowledge in \eprompt. Finally, we test the model on the target document type in a zero-shot fashion (Figure \ref{fig:intro}).

\subsection{Dual Prompting}
Given a serialized document represented as a sequence of tokens $\vx$ from the set of all documents $\mathbf{X}$, and a set of entities of interest $\mathbf{E} = \{\ve_1,\cdots, \ve_m\}$,
the goal is to let the model predict the corresponding label sequence $\vy$. In our query-based framework, we additionally define $\mathbf{Q} =  \{\vq_1,\cdots, \vq_m\}$ as the set of queries, where there is a bijection between $\mathbf{Q}$ and $\mathbf{E}$. The model takes an input tuple $(\vq_i, \vx)$ and predicts the conditional output $\vy_{\vq_i}$ (see BOISE prediction in Figure~\ref{fig:overview} for example). $\vy_{\vq_i}$ defines the token spans of the given query $\vq$ with 5 classes (\ie, BOISE).

To encode entity information into the query, we can use the entity name as query, \ie, $\vq_i = \ve_i$. We denote by $t$ the tokenizer, $f_\theta$ the input embedding layer and $p_\phi$ the rest of the language model. 
Then we can get the token-wise BOISE prediction:
\begin{align} \label{eq:entity_only}
    \begin{split}
       \hat{\vy}_{\vq_i}=  p_{\phi} ( f_\theta([t(\ve_i); t(\vx)])),
    \end{split}
\end{align}
where ``$[\cdot\ ;\ \cdot]$'' is the concatenation operation along the token length dimension. Note that although $\ve$ itself is not learnable, we can still learn its embedding $f_\theta(t(\ve))$ by optimizing $\theta$.

We name the query directly generated from the entity name as \eprompt.
However, in \methodabbr, our novel pre-training stage requires learning from large amount of webpages, which contain diverse categories of schema. Therefore, the model naturally requires more informative queries that also encodes the schema information. To this end, we propose the \dprompt to capture schema information. In pre-training, we can generate \dprompt in a similar way as we do for \eprompt, please see Section~\ref{sec:pretraining} for more details. During fine-tuning, the schema for form-like documents is often very different from that of webpages. Thus, we let the model learn the schema representation directly from the data, so that it can align well to the \dprompt{s} used in pre-training. We denote \dprompt by $\vs$, learnable vectors in the token embedding space during fine-tuning to capture schema information implicitly from the data.
According to the assumption in Section~\ref{sec:problem_formulation}, the documents we used in fine-tuning includes the target entities of interest. Intuitively, the schema information from fine-tuning documents should be transferrable to the target test document type. So we directly reuse the learned \dprompt when testing the target documents. In this case, $\vq_i = (\vs, \ve_i)$ and the prediction becomes:
\vspace{-2mm}
\begin{equation} \label{eq:bi_prompt}
    \hat{\vy}_{\vq_i} = p_{\phi} ( [\vs; f_\theta([t(\ve_i); t(\vx)])]).
\end{equation}
Finally, the model is trained with the objective:
\vspace{-.3cm}
\begin{align} \label{eq:ft-objective} 
     \min_{\theta, \phi, \vs}\ \sum_{\vx} \sum_{i=1}^{m} \mathcal{L} \left( \vy_{\vq_i}, \hat{\vy}_{\vq_i} \right), \vspace{-.3cm}
\end{align}
where $\mathcal{L}$ is the cross-entropy loss.

\subsection{\pagemage: Webpage-based Pre-training}
\label{sec:pretraining}
Distinct from recent work that focuses on multimodal pre-training, our proposed pre-training approach provides a new perspective with two core ideas: (1) Aligning pre-training and fine-tuning objectives. (2) Utilizing easy-accessible and informative webpages.

Recall that, in the fine-tuning stage, \methodabbr is trained with a moderately sized set of queries composed of \eprompt{s} generated from human-annotated entities and a learnable \dprompt that encodes schema information. It is reasonable to believe that if we can pre-train the model with an extremely large set of queries composed of different \eprompt and \dprompt generated from weakly-annotated documents, the model will perform better than the Masked Language Model (MLM)~\cite{devlin2018bert} pre-training alone. Here, we present our simple webpage-based pre-training technique as well as our data collection recipes to enpower the pre-training. 

\textbf{Dual prompting based pre-training.} We directly extract schema and entity information from rich HTML structure of various webpages and use them to generate the \dprompt and the
\eprompt, respectively. With a slight abuse of notation, we denote \dprompt by $\Tilde{\vs}$, where $\Tilde{\cdot}$ indicates that \dprompt is no longer a learnable parameter. Different from the fine-tuning stage with a single set of entities under a unified schema, we can group the webpages by schema: $\{(\Tilde{\vs}_1, \mathbf{E}_1, \mathbf{X}_1), \cdots, (\Tilde{\vs}_n, \mathbf{E}_n, \mathbf{X}_n )\}$, where each schema $\Tilde{\vs}_j$ corresponds to a set of entities $\mathbf{E}_j$ and a set of webpages $\mathbf{X}_j$. Similarly, the model takes the query-document tuple $(\vq_{ij}, \vx)$, where $\vq_{ji} = (\Tilde{\vs}_j, \ve_{ji})$, $\ve_{ji} \in \mathbf{E}_j$ and $\vx \in \mathbf{X}_j$, and outputs the following conditional prediction $\hat{\vy}_{\vq_{ji}}$:
\begin{equation} \label{eq:bi_prompt_pre}
    \hat{\vy}_{\vq_{ji}} = p_{\phi} ( f_\theta([t(\Tilde{\vs}_i); t(\ve_j); t(\vx)])).
\end{equation}
Equation~\ref{eq:bi_prompt_pre} is analogous to~\eqref{eq:bi_prompt}, however, $\vs_i$ here is directly sourced from webpage data, which makes it different from the learnable $\vs$ in~\eqref{eq:bi_prompt}.
Then we have the following pre-training objective:
\vspace{-3mm}
\begin{align} \label{eq:pre-objective}
     \min_{\theta, \phi}\  \sum_{j=1}^{n} \sum_{\vx \in \mathbf{X}_j} \sum_{i=1}^{|\mathbf{E}_j|} \mathcal{L} \left( \vy_{\vq_{ji}}, \hat{\vy}_{\vq_{ji}} \right).
\end{align}
The pre-training format is highly aligned with the fine-tuning in \eqref{eq:ft-objective}, so the model learns consistently during both stages to make query-conditional predictions.

\begin{figure}[t]
\begin{minted}[fontsize=\small]{html}
http://www.example.com
<div class=”product”>
<span id=”name”>Bath Mat</span>
<span id=”price”>$13.99</span>
</div>
\end{minted}
\vspace{-.5cm}
\caption{An example of HTML snippets, with two entities \textit{product/name} and \textit{product/price}.}
\label{figure:Webpage_example}
\end{figure}

\textbf{Data collection recipe.} 
How to extract schema and entity information from any webpage is another contribution of this paper. 
Consider the HTML snippets from Figure~\ref{figure:Webpage_example}.
First, it naturally contains two entities, and the combination of HTML tags defines what the entity is about, or its ``entity type''. Therefore, we have "Bath Mat" is of entity \textit{product/name}, and ``\$13.99'' is of entity \textit{product/price}. Second, the schema of the webpage = \{\textit{product/name}, \textit{product/price}\}, and the schema is usually shared by a series of similar webpages under the same domain.
Therefore, we can extract the domain name ``\textit{www.example.com}'' as the schema information.
Both the schema information and entity types generated from webpages are then respectively encoded by our \biprompt mechanism.
In practice, the schema and entity information automatically generated from webpages are often noisy. However, in the experiments, our model is still able to learn structured information from noisy queries and obtain significantly better entity extraction performance on the target form-like documents.
Moreover, in order to represent webpages in a manner that generalizes to form-like documents, the webpage representation consists only of the visible text tokens and corresponding $x/y$ coordinates.\footnote{Visible text and coordinates are generated by rendering each page with Headless Chrome: https://developer.chrome.com/blog/headless-chrome/.}

\begin{table}[t!] 
\begin{center}
\scalebox{0.72}{
\begin{tabular}{l|c|c|c}
\toprule 
 \bf Task & \bf Pre-training & \bf Train & \bf Test \\
\midrule
 Zero-shot & \multirow{2}{*}{\pagemage} & FUNSD & XFUND   \\
 Zero-shot &  & Inventory & Payment  \\
 \midrule
 Few-shot & \pagemage + Inventory & Payment X-shot & Payment  \\
 \bottomrule
\end{tabular}
}
\vspace{-.1cm}
\caption{Experiment design of two zero-shot transfer learning and one few-shot learning tasks.} %
\label{table:experiment_design}
\end{center}
\end{table}

\begin{table}[t!] 
\begin{center}
\scalebox{0.75}{
\begin{tabular}{l|r|r|r|r}
\toprule 
 \bf Dataset & \bf Schema & \bf Lang. & \bf Entity & \bf Example \\
\midrule
 FUNSD & 1 & 1 & 4 & 199 \\
 XFUND & 1 & 7 & 4 & 1393 \\
 Payment & 1 & 1 & 7 & 10k \\
 \midrule
 Inventory & 1 & 1 & 7 or 28 & 24k \\ \midrule
 \pagemage & 87K  & 1 & 2.6M & 1.2M \\
 \pagemage-ML & 113K & >50 & 11.5M & 13M \\
 \bottomrule
\end{tabular}
}
\vspace{-.3cm}
\caption{Detailed statistics of used datasets.} \vspace{-.5cm}
\label{table:dataset_stats}
\end{center}
\end{table}

\begin{table*}[t!]
    \centering
    \scalebox{0.70}{
    \begin{tabular}{l|cc|cc|c|ccccccc|c}
    \toprule
    \multirow{2}{*}{\textbf{Method}} & \multicolumn{2}{c|}{\textbf{Pre-training}} & \multirow{2}{*}{\textbf{Image}} & \bf \# Layers & \multirow{2}{*}{\textbf{FUNSD}} & \multicolumn{8}{c}{\textbf{XFUND}} \\
     & \bf Method & \bf Size & & \bf (Model Size) & & \bf ZH & \bf JA & \bf ES & \bf FR & \bf IT & \bf DE & \bf PT & \bf Avg.\\
    \midrule
    XLM-RoBERTa & MLM & 2.5TB &  & 12L(270M) & 66.70 & 41.44& 30.23 & 30.55 & 37.10 & 27.67 & 28.86 & 39.36 & 38.24 \\
    InfoXLM & MLM & 2.5TB & & 12L(270M) & 68.52 & 44.08 & 36.03 & 31.02 & 40.21 & 28.8 & 35.87 & 45.02 & 41.19 \\
    LayoutXLM & MLM & 30M & \checkmark & 12L(345M) & 79.40 & 60.19 & 47.15 & 45.65 & 57.57 & 48.46 & 52.52 & 53.90 & 55.61 \\ \midrule
    XLM-RoBERTa & MLM & 2.5TB &  & 24L(550M)& 70.74 & 52.05 & 39.39 & 36.27 & 46.72 & 33.98 & 41.80 & 49.97 & 46.37 \\
    InfoXLM & MLM & 2.5TB & & 24L(550M) & 73.25 & 55.36 & 41.32 & 36.89 & 49.09 & 35.98 & 43.63 & 51.26 & 48.35 \\
    LayoutXLM & MLM & 30M & \checkmark & 24L(625M) & \bf 82.25 & \bf 68.96 & 51.90 & 49.76 & 61.35 & 55.17 & 59.05 & 60.77 &  61.15 \\ \midrule
    \methodabbr & \pagemageabbr & 1.2M &  & 6L(82M) & 75.92 & 59.30 & 44.43 & 47.75 & 67.63 & 55.23 & 70.75 & 62.79 & 58.27 \\
    \methodabbr & \pagemageabbr-ML & 13M &  & 12L(185M) & 80.84  &  67.68  &  \bf 53.30 & \bf 55.80 & \bf 72.91 & \bf 67.30 & \bf 74.25 & \bf 69.30 & \bf 65.79 \\
    \bottomrule
    \end{tabular}
    }
    \vspace{-.2cm}
    \caption{Comparison between \methodabbr and competing methods on FUNSD-XFUND zero-shot benchmark. %
    } %
    \label{table:xfund-zs}
\end{table*}

\begin{table*}[t!]
    \small
    \centering
    \scalebox{0.85}{
    \begin{tabular}{l|cc|c|c|cc}
    \toprule
    \multirow{2}{*}{\textbf{Method}} & \multicolumn{2}{c|}{\textbf{Pre-training}} & \multirow{2}{*}{\textbf{Model Size}} & \bf Supervised & \multicolumn{2}{c}{\textbf{Source $\rightarrow$ Target}} \\
        & \bf Method & \bf Size & &  \bf Upper-bound & \bf I7 $\rightarrow$ Payment & \bf I28 $\rightarrow$ Payment \\
    \midrule
    ETC+RichAtt & MLM & 0.7M & 83M & 94.33 & 81.54 & 78.37 \\ 
    FormNet & MLM & 7M & 82M & \bf 95.70 & 79.77  & 77.88 \\ 
    FormNet & MLM & 7M & 157M & 95.61 & 86.04 & 82.42 \\ \midrule
    \methodabbr & \pagemageabbr & 1.2M & 82M & 94.60 & \bf 88.15 & \bf 89.23 \\
    \bottomrule
    \end{tabular}
    }
    \vspace{-.2cm}
    \caption{Comparison between \methodabbr and previous state-of-the-arts on Inventory-Payment zero-shot benchmark. I-7 and I-28 are abbreviations of Inventory-7 and Inventory-28, respectively. \methodabbr has much better generalization ability indicated by its stronger zero-shot performance and smaller gap with its supervised upper-bound (trained and tested both on Payment).} \vspace{-.2cm}
    \label{table:payment-zs}
\end{table*}

\section{Experiments}
\subsection{Datasets and Experiment Design}
We use 3 publicly available datasets and 2 in-house datasets that we collected to design and conduct extensive experiments to validate our method. Table \ref{table:dataset_stats} summaries the datasets.

\textbf{FUNSD}~\citep{jaume2019funsd} is a form understanding benchmark consisting of 199 annotated forms in English with %
4-entity types: \texttt{header}, \texttt{question}, \texttt{answer}, and \texttt{other}.

\textbf{XFUND}~\citep{xu2021layoutxlm} is a multilingual form understanding benchmark by extending the FUNSD dataset. The XFUND benchmark has 7 different languages with 1,393 fully annotated forms, where each language includes 199 forms with the same set of 4 entity types as FUNSD.

\textbf{Payment}~\citep{majumder2020representation} consists of around 10K documents and 7 entity types from human annotators. The corpus is collected from different vendors with various layout templates. In the few-shot learning experiments, we create multiple subsets by randomly subsampling documents from its training set.

\textbf{Inventory} is collected by us that contains inventory-related purchase documents in English (e.g., utility bills), containing a few document types different from the Payment dataset.
We prepare the dataset consists of $\sim$~24k documents in two annotated versions. The first version, \textit{Inventory-7}, is annotated at word-level with the same 7 entity types from Payment. The second version, \textit{Inventory-28}, is annotated at word-level with 21 additional entities types, including common entity types such as \texttt{shipping address}, \texttt{supplier name}. 

\textbf{\pagemage} is collected by us from publicly available webpages in English from the Internet with the acquisition procedure stated in Section~\ref{sec:pretraining}.

\textbf{\pagemage-ML} is a multilingual (ML) version of \pagemage with more than $50$ languages (at $99\%$ percentile). We collect the dataset to validate the effectiveness for multilingual pre-training for zero-shot generalizability across different languages.

\subsection{Experimental Details} \label{sec:exp_details}
We use the BERT-multilingual vocabulary~\cite{devlin2018bert} to tokenize the serialized OCR words. We have two variants of \methodabbr: a 6-layer ETC with 512 hidden size and 8 attention heads and a 12-layer ETC with 768 hidden size and 12 attention heads. For both \dprompt and \eprompt generated from dataset annotations, we use a maximum token length of 32 with zero padding. For learnable \dprompt used in the fine-tuning stage, we treat its token length as a hyperparameter to search.

Our method uses the proposed \pagemage pre-training approach on the 2 large-scale webpage-based datasets. Other comparing methods use MLM pre-training with the corresponding datasets mentioned in their papers, including $\sim$0.7k unlabeled form documents for ETC+RichAtt~\citep{ainslie2020etc}, IIT-CDIP dataset~\citep{lewis2006building} with 7M documents for FormNet~\cite{lee2022formnet}, 30M multilingual documents for LayoutXLM~\cite{xu2021layoutxlm}, and 2.5TB multilingual CommonCrawl data for XLM-RoBERTa~\cite{conneau2019unsupervised} and InfoXLM~\cite{chi2020infoxlm}.

We use mirco-F1 to evaluate the performance on XFUND related expeirments, following~\citet{xu2021layoutxlm}; macro-F1 to evaluate Payment related experiments, following~\citet{majumder2020representation, lee2022formnet}. We report the mean of best experiment results across three runs with difference seeds. Please see Appendix~\ref{app:exp_info} for more experimental details.

\begin{table}[t!]
    \small
    \centering
    \scalebox{0.85}{
    \begin{tabular}{c|cc|cc}
    \toprule
     \multirow{2}{*}{\textbf{Pre-training}} & \multirow{2}{*}{\textbf{E-P}} & \multirow{2}{*}{\textbf{S-P}} & \multicolumn{2}{c}{\textbf{Source $\rightarrow$ Target}} \\
     & & &  \bf I7 $\rightarrow$ Payment & \bf I28 $\rightarrow$ Payment \\
    \midrule
     MLM & \checkmark & & 81.95 & 85.71 \\
     \pagemage & \checkmark &  & 85.33 & 87.41 \\
     MLM & \checkmark & \checkmark & 84.91 & 86.41 \\
     \pagemage  & \checkmark & \checkmark & \bf 88.15 & \bf 89.23 \\
    \bottomrule
    \end{tabular}
    }
    \caption{Ablation study of \methodabbr on Inventory-Payment benchmark. E-P and S-P are abbreviations of \eprompt and \dprompt, respectively.} \vspace{-.3cm}
    \label{table:ablation}
\end{table}

\subsection{Zero-shot Transfer Learning Results}

\begin{figure*}[t]
    \centering
    \includegraphics[width=.95\textwidth]{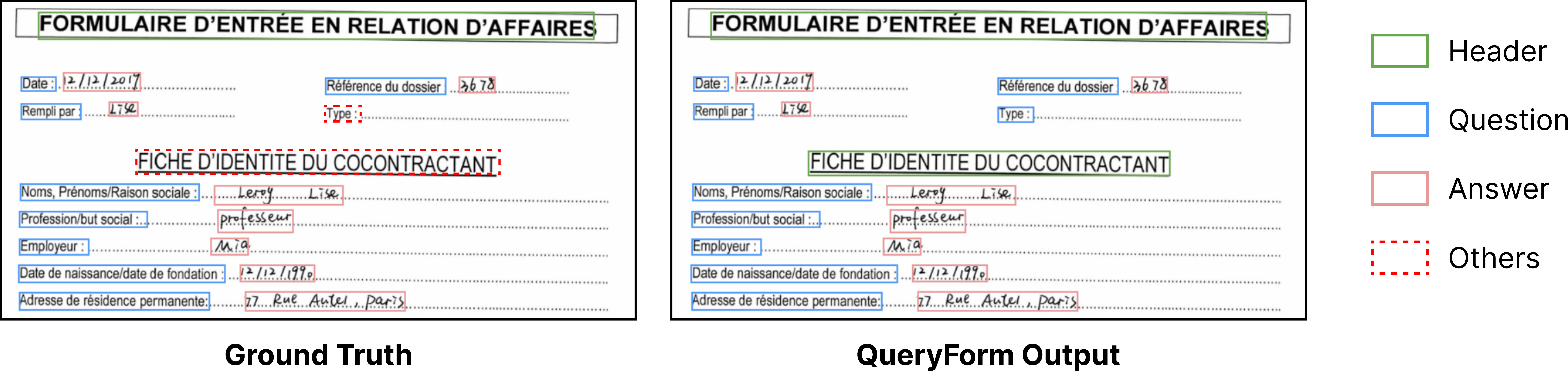}
    \vspace{-.2cm}
    \caption{Visualization example from XFUND (French). \methodabbr labels entities with ambiguous ``others'' annotation from ground truth as one of the other three entity types with concrete meanings.}%
    \label{fig:visualization} \vspace{-.1cm}
\end{figure*}

\begin{figure}[t]
\centering
\centering
\includegraphics[width=.48\textwidth]{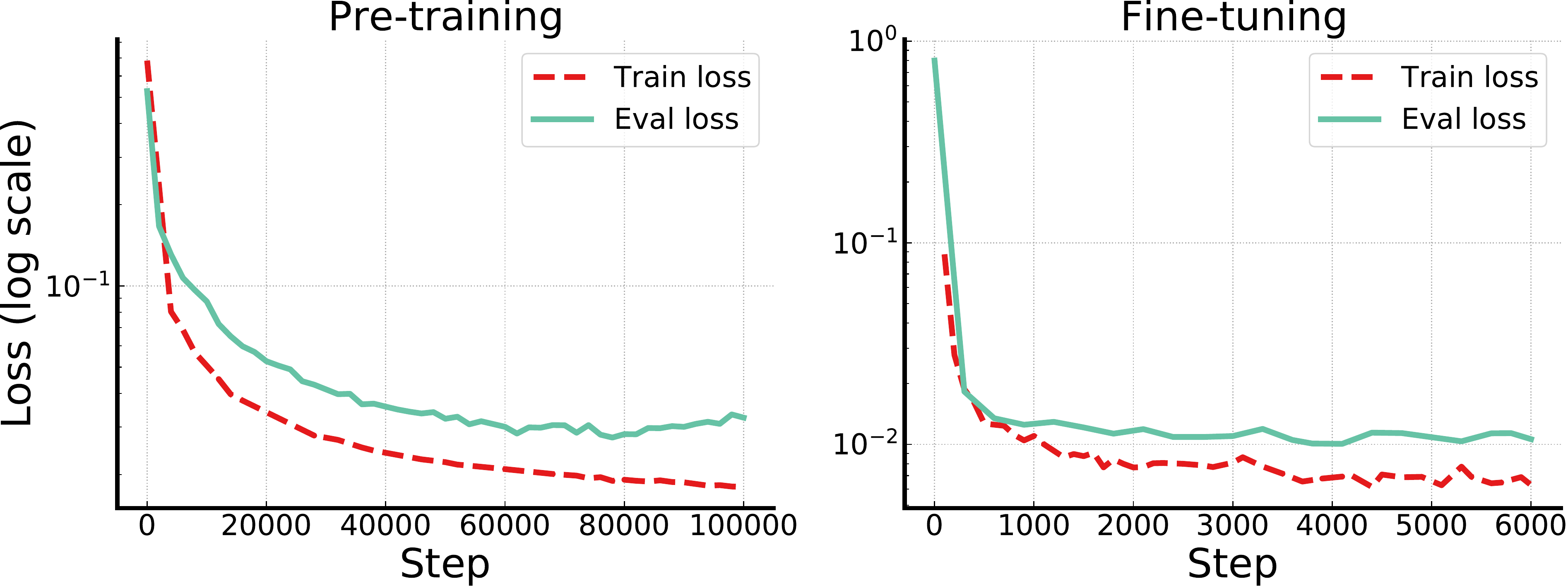}
\vspace{-.5cm}
\captionof{figure}{Loss visualization of pre-training on \pagemage (Left) and fine-tuning on Inventory (Right).}
\vspace{-.4cm}
\label{fig:train_loss}
\end{figure}

To evaluate \methodabbr, we introduce two zero-shot transfer learning tasks and one few-shot learning task, as shown in Table~\ref{table:experiment_design}. We follow the official train-test split for all public available datasets by default, unless specified explicitly. 
For zero-shot on Payment test set, we pre-train on \pagemage and fine-tune on Inventory. 

\noindent \textbf{FUNSD-XFUND.} %
In Table~\ref{table:xfund-zs}, we compare \methodabbr with recent zero-shot transfer learning methods, include XLM-RoBERTa~\citep{conneau2019unsupervised}, InfoXLM~\cite{chi2020infoxlm}, and the current state-of-the-art LayoutXLM~\cite{xu2021layoutxlm}, which are all MLM pre-trained on multilingual text or document datasets of different sizes (details in Section~\ref{sec:exp_details}).
\methodabbr outperforms all comparing methods even with much smaller model size and no image modality. In particular, when we pre-train \methodabbr on the multilingual \pagemage-ML, \methodabbr obtains a significant boost on all languages.
Although XLM-RoBERTa and InfoXLM are MLM pre-trained on 2.5TB multilingual data, and LayoutXLM especially collected 30M visually rich documents for MLM pre-training, the stronger transferability of \methodabbr for zero-shot DEE indicates that our pre-training method is more effective than MLM for this specific task. Since the 12 layer (185M) \methodabbr has outperformed previous state-of-the-arts by a large margin with a much smaller model size, we leave further scaling up as future work.

\begin{table}[t!]
\small
    \centering
    \scalebox{0.85}{
    \begin{tabular}{l|c|c}
    \toprule
    {\textbf{Method}} & {\textbf{Payment 1-shot}}  & {\textbf{Payment 10-shot}}\\
    \midrule
    ETC+RichAtt\textsuperscript{2} & 59.62 & 88.21 \\
    FormNet\textsuperscript{2} (157M) & 55.67 & 86.25 \\
    \methodabbr & \textbf{89.26} & \textbf{90.53} \\
    \bottomrule
    \end{tabular}
    }
    \caption{Comparison between \methodabbr and comparing methods further fine-tuned on  few-shot Payment training sets.}
    \vspace{-3mm}%
    \label{table:payment-fs}
\end{table}

\noindent \textbf{Inventory-Payment.} 
Table~\ref{table:payment-zs} shows the zero-shot transfer learning result on the Inventory-Payment benchmark. 
We compare \methodabbr against the current state-of-the-art on Payment, FormNet~\cite{lee2022formnet}, and our baseline ETC+RichAtt (see Section~\ref{sec:arch_design}). \methodabbr outperforms competing methods by a significant margin. Although FormNet obtains the best supervised upper-bound result on Payment, the lower zero-shot results indicate that knowledge transfer from different types of documents is still very challenging. 

\methodabbr is expected to take advantage of larger number of queries though they are less relevant.
To validate, we compare fine-tuning datasets with 7 and 28 annotated entities. As can be seen, supervised methods like FormNet and ETC+RichAtt suffer performance drop when seeing additional entities not existed in target dataset, while \methodabbr gains further performance improvement.

\noindent \textbf{Ablation study.} We conduct ablation study of \methodabbr. From the results in Table~\ref{table:ablation}, we can see that both our \biprompt strategy and \pagemage pre-training contribute to the zero-shot F1 score individually, and synergistically improve the performance when working together.

\subsection{Few-shot Learning on Payment} 
In practice, %
it is reasonable to believe that a few annotated document from the target document type can make models quickly adapt.
Therefore, we design a few-shot learning step based on the best performing model obtained on the Inventory dataset.

Table~\ref{table:payment-fs} shows the 1- and 10-shot results on Payment. To make sure the training is stable on low data regime and comparison is fair, we conduct hyperparameter search (\eg, learning rates, \# of freezing layers) for all methods and select the best performing ones to present. When fine-tuning on the extreme Payment 1-shot, both FormNet and ETC-RichAtt overfit the single document from Payment severely, while \methodabbr maintains high performance\footnote{For compared methods trained by us, although 10-shot leads to improvement, 1-shot degrades with the same parameter-search methods used for \methodabbr. More advanced training strategies are not considered here.}.%
When extending to 10-shot, all methods improves, and \methodabbr still perform the best. {Although FormNet is state-of-the-art in supervised learning setting, it underperforms other methods on low data regime. We hypothesize GCN requires more data to learn layout features.}

\subsection{Result analysis}

\noindent \textbf{Prediction visualization.} Figure~\ref{fig:visualization} demonstrates an example output of \methodabbr. %
\methodabbr infers entities that are annotated as ``others'' in ground truth as one of the other three entity types with concrete meanings. For example, ``Type'' is a question in the form, however, without corresponding answer. Although human annotators might find it ambiguous and mark it as ``others'', \methodabbr successfully recognize it as a ``question''.

\noindent \textbf{Loss visualization.} %
Figure~\ref{fig:train_loss} shows the loss curves of pre-training on \pagemage (Left) and fine-tuning on Inventory (Right). According to the pre-training loss curve, we observe that the loss converges well despite the fact that the weak supervision extracted from webpages is often noisy. Moreover, according to the fine-tuning curve, we observe that the loss converges very fast, thanks to the knowledge learned during pre-training. The observations indicate that our framework successfully extracts useful information from the weak supervision and leverages the learned information to facilitate fine-tuning on form-like documents.

\section{Conclusion}
This paper presents \methodabbr, a novel framework to address the challenging zero-shot document entity extraction problem. The \biprompt design in \methodabbr offers a refreshing view to unify the pre-training and fine-tuning objectives, allowing us to leverage large-scale form-like webpages with HTML tags as weak annotations.
\methodabbr sets new state-of-the-art results on multiple zero-shot DEE benchmarks. 
We believe \methodabbr serves as a flexible framework for document understanding tasks, and multiple interesting directions could be further explored within the framework, such as prompt design, richer pre-training sources, etc.

\section*{Ethical and Broader Impact} \label{sec:broader_impact}
We have read the ACL Code of Ethics, and ensure that our work is conformant to this code. As a novel framework for zero-shot document entity extraction, \methodabbr has a great potential to boost the performance of existing DEE systems. However, we would still like to discuss the limitations and risks to avoid any misuse of \methodabbr.

Although our proposed \pagemage pre-training approach can effectively achieve knowledge transfer from publicly available webpages to form-like documents, it inevitably carries the bias and fairness problems~\cite{mehrabi2021survey} to the downstream task. Therefore, in real-world applications, we should have more strict rules to filter and clean up the webpages, and thoroughly check the bias and fairness issues of the pre-trained model.

\section*{Limitations}
In addition to the bias and fairness concerns that we discussed in the Ethical and Broader Impact section, we discuss the possible limitations of our method in this section.

As a query-based DEE framework, QueryForm may be prone to specific prompting based adversarial attacks~\cite{xu2022exploring}, which may further pose potential security concerns for safety-critical documents. Thus, it is important to test the robustness of QueryForm against adversarial attacks and design defense schemes to further strengthen our method in the future.

Our work focuses on the closed-world setting that source documents include entities contained in the target documents, following~\cite{xu2021layoutxlm}, without further investigating the possible open-world~\cite{shu2018unseen} setting with unseen test entities. However, as a query-based framework that makes conditional prediction with no pre-defined set of entities, QueryForm actually supports the prediction of unseen entities at test time and we would like to leave it as an interesting future research direction.

\section*{Acknowledgements} 
We greatly thank Chun-Liang Li, Harr Chen, Evan Huang, Nan Hua for their valuable feedback.

\clearpage
\appendix
\section{Additional Dataset Information}
\textbf{Licensing information.} We provide the licensing information of publicly available dataset as follows:
\begin{itemize}
    \item \textbf{FUNSD} is licensed under the license specified in~\url{https://guillaumejaume.github.io/FUNSD/work/}, where the use of the FUNSD Dataset is solely for non-commercial, research and educational purposes.
    \item \textbf{XFUND} is licensed under the Attribution-NonCommercial-ShareAlike 4.0 International (CC BY-NC-SA 4.0) license.
    \item \textbf{Payment} has no licensing information, however, we use it accordingly following~\citet{majumder2020representation,lee2022formnet}.
\end{itemize}
The use of existing artifacts is consistent with their intended use for research and educational purposes.

\noindent \textbf{Collection of in-house datasets.} For datasets that we collect, both Inventory and \pagemage are sufficiently anonymized to remove personal information. Moreover, our internal system filters offensive content when collecting public available webpages. The possible limitation of our current data collection methodology for \pagemage is that we do not enforce strict check on the bias and fairness issues, as discussed in Section~\ref{sec:broader_impact}. Our method mainly leverages layout and HTML-related information from webpages instead of the semantic content. No images are used. Additionally, it is useful to further improve our data collection method in the future to avoid potential bias and fairness issues.

\section{Additional Experiment Information} \label{app:exp_info}
For \pagemage pre-training, we use Adam optimizer with a batch size of 512. We set the learning rate to 0.0002 with a warm-up proportion of 0.01 and a linear learning rate decay of 100k steps.
We fine-tune all models using the Adam optimizer with a batch size of 128 and a learning rate of 0.0004. No warm-up or learning rate decay is used. Pre-training takes approximately 48 hours on 8x8 TPU v3. Fine-tuning takes approximately 4 hours on the largest corpus, Inventory, on both TPUs and Tesla V100 GPUs. We implement our method based on the same codebase used in~\citet{lee2022formnet}.

\end{document}